\documentclass[a4paper,fleqn]{cas-dc}

\usepackage[numbers]{natbib}
\usepackage{hyperref}
\usepackage{color}
\usepackage{ulem}
\usepackage{subfig}
\usepackage{longtable}
\usepackage{supertabular,booktabs}
\usepackage{makecell}
\usepackage{multicol,lipsum}
\usepackage{graphicx}
\usepackage{url}
\usepackage{hyphenat}
\def\tsc#1{\csdef{#1}{\textsc{\lowercase{#1}}\xspace}}
\tsc{WGM}
\tsc{QE}
\tsc{EP}
\tsc{PMS}
\tsc{BEC}
\tsc{DE}
\begin{document}

\let\WriteBookmarks\relax
\def\floatpagepagefraction{1}
\def\textpagefraction{.001}
\shorttitle{A Rule-Based Model for Victim Prediction}
%\shortauthors{CV Radhakrishnan et~al.}

\title [mode = title]{A Rule-Based Model for Victim Prediction}                      
%\tnotemark[1,2]

%\tnotetext[1]{This document is the results of the research
%   project funded by the National Science Foundation.}

%\tnotetext[2]{The second title footnote which is a longer text matter
%   to fill through the whole text width and overflow into
%   another line in the footnotes area of the first page.}

\author[1]{Murat Ozer}[ orcid=0000-0002-7582-1298]%[type=editor,
                     %   auid=000,bioid=1,
                     %   prefix=Sir,
                    %    role=Researcher,
                    %    orcid=0000-0001-7511-2910]
\cormark[1]
\ead{ozermm@ucmail.uc.edu}
%\ead[url]{www.cvr.cc, cvr@sayahna.org}

%\credit{Conceptualization of this study, Methodology, Software}

\address[1]{School of Information Technology, University of Cincinnati, Ohio, United States}

\author[1]{Nelly Elsayed}[orcid=0000-0003-0082-1450]
%\cormark[2]

\author[1]{Said Varlioglu}[%
   orcid=0000-0003-4013-6386
   ]
   
\author[1]{Chengcheng Li}[%
orcid= 0000-0003-2945-0462
]

\author[2]{Niyazi Ekici}[%
orcid= 0000-0001-9565-4821
]

\address[2]{Law Enforcement and Justice Administration, Western Illinois University, Illinois, United States}

%\ead{cvr3@sayahna.org}
%\ead[URL]{www.sayahna.org}

%\credit{Data curation, Writing - Original draft preparation}

%\address[2]{Sayahna Foundation, Jagathy, Trivandrum 695014, India}

%\author%
%[1,3]
%{Rishi T.}
%\cormark[2]
%\fnmark[1,3]
%\ead{rishi@stmdocs.in}
%\ead[URL]{www.stmdocs.in}

%\address[3]{STM Document Engineering Pvt Ltd., Mepukada,
%    Malayinkil, Trivandrum 695571, India}

%\cortext[cor1]{Corresponding author} commented out on 3/6/2022------

%\cortext[cor2]{Principal corresponding author}
%\fntext[fn1]{This is the first author footnote. but is common to third
%  author as well.}
%\fntext[fn2]{Another author footnote, this is a very long footnote and
%  it should be a really long footnote. But this footnote is not yet
%  sufficiently long enough to make two lines of footnote text.}

%\nonumnote{This note has no numbers. In this work we demonstrate $a_b$
%  the formation Y\_1 of a new type of polariton on the interface
%  between a cuprous oxide slab and a polystyrene micro-sphere placed
%  on the slab.
%  }

\begin{abstract}
The present study proposes a novel automated model, called Vulnerability Index for Population at Risk (VIPAR) scores, to identify rare populations for their future shooting victimizations. Likewise, the focused deterrence approach identifies vulnerable individuals and offers certain treatments (e.g., outreach services) to prevent violence in communities. Our rule-based engine model is the first AI-based model for victim prediction purposes. The model merit is the usage of criminology studies to construct the rule-based engine to predict victims. This paper aims to compare the list of focused deterrence strategy with the VIPAR score list regarding their predictive power for the future shooting victimizations. Drawing on the criminological studies, this study uses age, past criminal history, and peer influence as the main predictors of future violence. Network graph analysis is employed to measure the influence of peers on the outcome variable. The proposed model also uses logistic regression analysis to verify the variable selections in the model. Following the analytical process, the current research creates an automated model (VIPAR scores) to predict vulnerable populations for their future shooting involvements. Our empirical results show that VIPAR scores predict 25.8\% of future shooting victims and 32.2\% of future shooting suspects, whereas the focused deterrence list predicts 13\% of future shooting victims and 9.4\% of future shooting suspects. The proposed model outperforms the intelligence list of focused deterrence policies in predicting the future fatal and non-fatal shootings. Furthermore, this paper discusses the concerns about the presumption of innocence right. 

\end{abstract}

%\includegraphics{figs/grabs.pdf}

%DELETED by Said
%\\begin{highlights}
%\\item Predicting a shooting attack victim using a rule-based system that established on criminology theory.
%\\item The first artificial intelligent-based victim prediction model.
%\\item The prediction results outperforms the existing stat-of-the-art victim prediction models.
%\\end{highlights}

\begin{keywords}
Rule-based system\sep Network graph analysis\sep Violent victimization\sep Victim prediction
\end{keywords}

\maketitle

\section{Introduction}\label{Sec:Intro}
Criminological theories have been studying the covariates of chronic offenders for long years since the early work of Sheldon and Eleanor Glueck~\cite{glueck1930500}. 
Subsequent studies found that age of onset~\cite{nagin1992stability, bartusch1997age, nagino1992onset, sampson1995crime, shannon1988criminal, farrington1986age, farrington1995effects,west1973becomes} seriousness of crime~\cite{nagin1995life} past criminal history~\cite{gottfredson1990general, nagin1991relationship} and delinquent peers~\cite{mcdermott2001same, paternoster1997multiple} are the main predictors of future delinquent behaviors as well as career criminals. Except for Gottfredson and Hirschi~\cite{gottfredson1990general}, researchers studying criminal careers suggest that desisting from crime is possible through enhancing social environments such as family structure and economic conditions. Therefore, the identification of chronic offenders is a crucial factor in the ability to implement a specific type of intervention. 

The studies mentioned above also revealed that chronic offenders account for less than five percent of the population, but they commit most of the overall crime~\cite{piquero2003criminal, shannon1988criminal, wolfgang1972delinquency}. Likewise, studies in the city of Cincinnati, Ohio, USA showed that a population of less than 0.05\% involved in the gang activity accounted for 75\% of all homicides and 50\% of the violent crimes~\cite{engel2013reducing}. Therefore, identifying career criminals have the most potential impact to reduce violent crime through certain types of interventions such as youth outreach services and pulling levers focused deterrence programs~\cite{braga2014deterring, kennedy1996youth, losel2012child, peterson2004gang}.

To date, there is no systemic approach to detect chronic offenders in the population to develop proactive approaches. Pulling levers strategy is the closest candidate in this realm, which echoes well-known findings of criminological theories that a small number of individuals, usually socially connected through co-offending networks, commit 48 to 90 percent of violent crimes across the United States~\cite{NGIC2011}. The offenses committed by these individuals are a behavioral byproduct of street norms that address violence as a means of solving problems and disrespect~\cite{kennedy1996youth}. More importantly, violence spreads with the influence of peers in these co-offending networks~\cite{akers1979lanza, shaw1942juvenile, warr1993parents, warr1991influence}. Therefore, directing resources on these individuals and group structure through social services and increased certainty of punishment gives promising results regarding reducing violence in cities~\cite{braga2014deterring, braga2013spillover, engel2013reducing}. Please note that we intensify our efforts to successfully understand the influence of co-offending networks on individuals. Criminological theories suggest that offenders and victims are often linked and routine interaction with offenders significantly increases the chance of future victimization \cite{cohen1979social, hindelang1978victims, mustaine2000comparing, sampson1990deviant}. Therefore, identifying co-offending networks is essential to prevent future victimizations. Focused deterrence policing is an example of this notion because this approach is based on the identification of a co-offending network to prevent future victimizations.

To implement a focused deterrence approach, law enforcement agencies organize intelligence-gathering sessions to identify criminals and groups who commit most of the violent crimes in cities. These sessions are usually done twice a year as it takes a considerable amount of time and is limited to the knowledge of attended field officers. However, street violence is incredibly dynamic, requiring almost real-time updates to capture vibrant street life, such as ongoing disagreements, disrespect between group members, and the emergence of new groups of violence. Focused deterrence strategy, on the other hand, captures one snapshot of the street life with the intelligence-gathering sessions, but intelligence quickly fades out with the new developments of the street dynamics. For this reason, the reduction of violence in the focused deterrence approach is often not self-sustaining over time~\cite{braga2012effects, corsaro2017assessing, tillyer2012beyond}.

\begin{figure*}
	\centering
	\includegraphics[width=13cm, height=2.8cm]{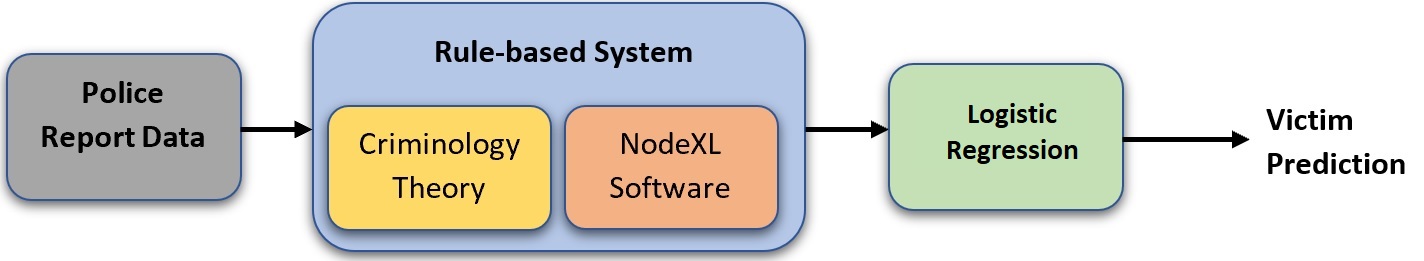}
	\caption{The proposed VIPAR model diagram.}
	\label{Fig:fig1}
\end{figure*}

Given this context, we proposed a novel rule-based system model that constructed under the criminological theory and offers a big data-based model that instantly captures street dynamics to evolve the list of chronic offenders who drive the violence in the city. The proposed rule-based model has a higher future violence prediction accuracy compared to the static chronic offender list of focused deterrence strategies. Hence, violence reduction can be realized at the optimum level. Therefore, the proposed model seeks an answer to the following questions: (1) to what extent law enforcement data allow researchers to predict future fatal and non-fatal shooting victims and suspects, (2) whether the proposed model better predicts the future shooting violence compared to the static list of chronic offenders, and (3) whether prioritizing individuals for their vulnerabilities to a crime lead to profiling concerns (e.g., racial, gender and place-based). The proposed model diagram is shown in Figure~\ref{Fig:fig1}. The next sections of this paper demonstrate a detailed explanation of the proposed model components followed by the empirical results and analysis.  

\section{Rule-Based System}
The rule-based system is an artificial intelligence technique to obtain significant information based on interpreting previous knowledge and experiences that have been already stored and assigned with different scores~\cite{davis1984origin,fox1990ai}. Usually, these sets of rules are assigned by a human expert. However, few systems are based on automatic rule inference design~\cite{fox1990ai}. One of the most popular rule-based systems is the expert system, the idea which was developed in the late 1940s\-early 1950s as a medical diagnostic machine~\cite{yanase2019systematic}. In the proposed model, we designed a ruled-based system to construct a risk score assignment mechanism based on the criminological theory. Thus, it enhances the overall predictive performance of the proposed model as it focuses only on the usage of appropriate attributes that have been proven as significant risk factors for the successful prediction of victims. The constructing of our proposed VIPAR rule-based system will be discussed in section~\ref{sec:VIPAR}

\section{Model Criminological Roots}

As previously stated in section~\ref{Sec:Intro}, criminological theories stress that age, past criminal history, and peer influence are the main predictors of future delinquent behavior. It is widely accepted in criminology that involvement in crime diminishes with age~\cite{farrington1986age}. In our proposed model, we used the criminological theories to construct our rule-based system rule sets and scores. 

In addition to this, life course theories suggest that individuals who start committing crime at an early age are more likely to continue committing a crime in the future~\cite{moffitt2017adolescence, moffitt2003life}. Likewise, past criminal behavior/history is one of the more robust predictors of future offending according to numerous studies/theories, including general theory of crime~\cite{gottfredson1990general, hanson2000should, langan2002recidivism, loeber1982stability, reiss1988co}.  Finally, the topic of peer influence attracted many criminologists for its ability to explain the disproportionate concentration of crime (e.g., social disorganization theory) and criminogenic behavior (e.g., differential association theory or social learning theories). 

In this context, law enforcement data easily allow researchers to extract age and past criminal history of individuals. Detecting peer influence; however, is time and labor intensive for many law enforcement agencies. For this reason, it is mostly disregarded in the analyses. Study findings, on the other hand, reveal that peer influence significantly predicts future violence. In this vein, Conway and McCord's~\cite{conway2002longitudinal} longitudinal study showed that offenders who committed their first co-offense with violent delinquent peers are more likely to commit violent crimes compared to those who were not exposed to violent offenders. In addition to the effect of peer influence on learning processes, Warr~\cite{warr1996organization} found that structural characteristics of co-offending networks\footnote{Warr (1996) describes structural characteristics as “delinquent groups”. Similarly, we term structural characteristics in the study as the number of violent individuals, the number of shootings, and the number of firearm-related incidents in each co-offending group.} influence individuals' behavior far beyond their characteristics/traits. Similarly, employing Add Health data, Haynie \cite{haynie2001delinquent} demonstrated the network characteristics of individuals (e.g., occupying a key role in a criminal offending network or involvement in a dense social network) influenced the outcome of individual propensities. 

Besides, Haynie's analyses~\cite{haynie2001delinquent} suggest the relationship between delinquency and peer association behaves differently in the context of network characteristics, which fundamentally demonstrates the more significant impact of peer influence over that of individual tendencies.

Recent studies, employing more sophisticated data and techniques, also found that co-offending networks explain gunshot victimization better than do: gender, race, or gang affiliation~\cite{papachristos2012social}. Papachristos et al.~\cite{papachristos2012social} employed police Field Intelligence Observation records to generate a network\footnote{Co-offending networks are generated by simply identifying two or more individuals involved in the same incident such as arrest, field interview, or victimization.}
for 238 known gang members until the second degree of friendship. Then the authors merged this network data with fatal and non-fatal shootings that occurred in 2008-2009. Further analysis revealed that closeness to a gunshot victim significantly increased the odds of subsequent gunshot victimization. Similarly, by studying the arrest data of a co-offending network, Papachristos, Wilderman, and Roberto~\cite{papachristos2015tragic} found that co-offending networks dramatically increase the likelihood of gunshot victimization, even more so than individual demographics or gang affiliation. Their study demonstrated that not only do one's immediate co-offending friends (direct exposure) increase the chance of victimization but also the friends of one's friends (indirect exposure) increase the likelihood of gunshot victimization.
In summary, studies on co-offending networks strongly emphasize that any violence prediction has to include peer influence and as well as the group structure itself (e.g., violent vs. non-violent co-offending groups). Therefore, the proposed model comprises measurements of co-offending networks during the estimation/prediction process. This paper explains the components of the proposed victim prediction model in detail and shows the success of the proposed model empirically compared to state-of-the-art victim prediction models. 

\section{Datasets}

In our empirical study, we employ six different datasets from the city of Cincinnati Police Department in the State of Ohio, United States. The first dataset is reported crimes (N= 176,660) from January 1, 2010, to December 31, 2014, which includes variables of the incident such as date, location, crime type, and modus operandi. The second dataset is suspect (N=33,480) and victim (N= 190,590) data of the reported crimes. This dataset includes suspect and victim demographics such as name, age, race, gender, and suspect-victim relationship. The third dataset is the arrest data (N= 122,542) for the same period that includes arrestee's demographics (race, sex, and date of birth), the location of arrest, and crime types.  The fourth dataset is the Field Interview Reports (N=228,796), which includes demographics of individuals as a result of traffic stops or pedestrian stops. The fifth dataset is the fatal and non-fatal shootings for the period of January 1, 2010, to December 31, 2015, that includes the demographics of shooting victims (N=2,511). The last dataset is a chronic offender list (N=3,215), which was compiled during the various intelligence gathering sessions of focused deterrence strategy applied in Cincinnati~\cite{engel2013reducing}. Cincinnati focused deterrence approach is known as the Cincinnati Initiative to Reduce Violence (CIRV). For this reason, this dataset is named as CIRV List that holds chronic offenders for their possible involvement of future shooting violence.

\section{Analytical Process}

As mentioned earlier, this paper aims to make a comparison between the CIRV List and the proposed model for predicting shooting violence. Therefore, we split the fatal and non-fatal shooting data into two different waves. During the model development, we used shooting victim data from January 1, 2010, to December 31, 2014 (N=2,034). We left aside 2015 fatal and non-fatal shooting victim data (N=477) to test how CIRV List and the model predict the future shooting victims. Likewise, we obtained 2015 known shooting suspects (N=149) to measure the prediction performance between CIRV List and our proposed model.We obtained known shooting suspect data by linking known suspects and arrestees to the victim data. 

Creating the Vulnerability Index: The proposed model aims to detect rare populations for their vulnerability to fatal and non-fatal shootings. For this reason, as of this point, the present study will use the Vulnerability Index for Population at Risk (VIPAR) scores interchangeably with the model. As discussed earlier, extracting age and criminal history from law enforcement data is relatively straightforward. Detecting the influence of peers in developing violence, however, needs time-intensive analysis. Therefore, any efficient violence prediction model should find ways to calculate the influence of co-offending networks automatically. Given this context, VIPAR rule-based system scores automate this process using various data mining steps, as explained below.

\begin{figure}
	\centering
	\includegraphics[width=0.6\columnwidth]{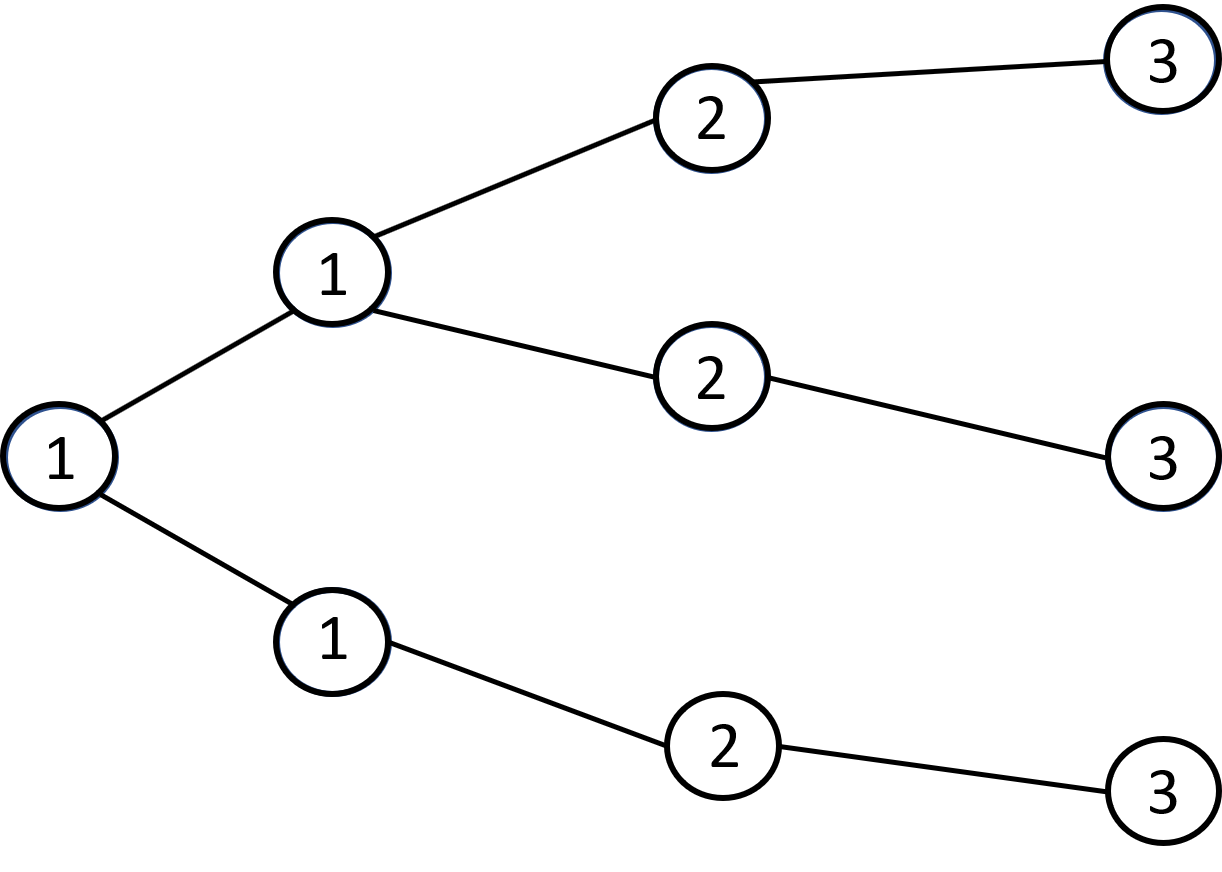}
	\caption{Creating Co-Offending Network}
	\label{Fig:fig2}
\end{figure}

For the first step, if two or more individuals were arrested, victimized, stopped (for field interview), or committed a crime together, the model assumes that those individuals are associated with each other by sharing the same event. The first step is called a first-degree co-offending network, which emphasizes the immediate friendships (in this case, the co-offending) based on a single event. This first-degree network can be expanded by finding the friends of friends of the first identified individuals\footnote{In the network graph theory, individuals are called nodes, and their relationships are called edges. The present study will keep the terms simple and will try to avoid using the jargon of network graph theory.}
by looking at the different events in which first degree friends involved with other individuals. Finally, the entire co-offending network can be expanded again by finding friends of friends of friends, which is called as a third-degree co-offending network. Figure~\ref{Fig:fig2} illustrates the first-degree co-offending network by assigning one (1) value to the circles. Likewise, values 2 and 3 represent the second and third-degree co-offending networks. 

Expanding co-offending networks until the third degree is necessary because criminology research suggests that not only one's immediate friends can increase the likelihood of violent crime involvement but also having a violent friend within a third-degree co-offending network, significantly increases the chance of violent offending~\cite{fujimoto2012network, papachristos2015tragic}. Therefore, VIPAR scores employ a co-offending network until the third degree.

By following the above analytical approach, the model generated a co-offending network, as shown in Table~\ref{table1}. Most of the relationships come from field interviews following suspect-victim relationships, victim relationships, and arrest/ suspect relationships. For instance, the Cincinnati Police Department (CPD) stopped 19,958 individuals between 2010 and 2014. These field interviews roughly accompanied by two individuals on average during the traffic or pedestrian stops. Even though the co-offending network seemingly includes 85,065 individuals, certain individuals repetitively involved in multiple criminal activities such as committing a crime and being a victim of a crime. For this reason, the number of unique individuals in the co-offending network is 55,454 when these duplicates were removed. From this perspective, the average number of relationships in the network is 2.42 (133961/ 55454).

\begin{table*}%[ht]
	\caption{Summary Statistics of Co-offending Network}
	\label{table1}
	\setlength\tabcolsep{1mm}
	\centering
	\begin{tabular}{L|C|C|C}
		\hline
		\textbf{ Source Data}&\textbf{\# of Individuals}&\textbf{\# of Relationships}&\textbf{Average Relationships}\\
		\hline
		Arrest/Suspect&10276&14362&1.40\\
		Field Interviews&19958&40657&2.04\\
		Suspect-Victim&39341&56517&1.44\\
		Victim&15490&22425&1.45\\
		\hline
		Total&15490&133961&1.57\\
		\hline
	\end{tabular}
\end{table*}

\section{VIPAR Measures}\label{sec:VIPAR}
The proposed VIPAR score model uses 21 variables and 25 different weights for constructing the rule-based system scheme. During the refinement of the model, certain variables were dropped from the model, such as ego density (i.e., number of immediate friends) and closeness to a CIRV List member at the second-degree co-offending network, due to their weak or no explanatory power for future crime prediction. As previously stated, the proposed model selects the appropriate rule from the system by following the proven thoughts of criminological theories. After the initial selection of the variables based on their availability in the police data, I tested them in the logistic regression model to see whether the selected variables significantly explain the likelihood of future fatal and non-fatal shootings. 

Likewise, each weight of the variable in the model is determined by looking at the magnitude of the relationship in the logistic regression model. Given this context, we classified the variables into three categories: personal variables, positional co-offending variables, and structural co-offending variables and tested their significance in the logistic regression model. 

\subsection{Personal Variables}
As shown in Table \ref{table2}, the personal variable category includes age, CIRV list membership, and past criminal history variables. Except for age, all other variables are measured in a dummy format. The percentage of CIRV members in the overall co-offending network is 11.9 (N=659). 

\begin{table*}%[ht]
	\caption{Descriptive Statistics of Personal Variables (N=55,454)}
	\label{table2}
	\setlength\tabcolsep{1mm}
	\centering
	\begin{tabular}{L|C|C}
		\hline
		&\textbf{Min - Max}&\textbf{Mean - \%}\\
		\hline
		Age&	13 - 80.9&	37.44 – 13.15(sd)\\
		CIRV List Member&	0 - 1&	11.90\%\\
		Recent  Violent Crime&	0 - 1&	4.10\%\\
		Recent Misdemeanor Crime (2 or more)&	0 - 1&	7.40\%\\
		Misdemeanor Crime (3 or more)&	0 - 1&	11.90\%\\
		Recent Firearm Involved Incident (arrest or victim)&	0 - 1&	5.30\%\\
		\hline
	\end{tabular}
\end{table*}

Note that, if a CIRV member is not involved in any criminal activity with others, that person will not be in the co-offending network. In this context, the current data suggest that 659 out of 3215 CIRV list members involved in crimes with accompanies. All the other variables in Table 2 capture the past criminal history of individuals from the police data.

\begin{table*}
	\caption{Personal Variables Predicting Future Fatal and Non
-
Fatal Shooting Victims}
	\label{table3}
	\setlength\tabcolsep{1mm}
	\centering
	\begin{tabular}{L|C|C|C|C}
		\hline
		&	\textbf{b}&	\textbf{S.E.}&	\textbf{p-value}&	\textbf{Exp(B)}\\
		\hline
		Age&	-0.043&	0.007&	0&	0.957\\
		CIRV List Member&	1.374&	0.218&	0&	3.951\\
		Recent Misdemeanor Crime (2 or more)&	0.649&	0.178&	0&	1.913\\
		Misdemeanor Crime (3 or more)&	1.465&	0.173&	0&	4.327\\
		Recent Firearm Involved Incident (arrest or victim)&	0.597&	0.185&	0.001&	1.817\\
		
		\hline
	\end{tabular}
\end{table*}

Table~\ref{table3} displays the logistic regression result of the future fatal and non-fatal shooting victims for the personal variables of the study. Although the odds-ratio (Exponent B) of the age variable seems to be small, age is the most influential variable in the equation. Note that, age is measured in decimals to reflect the precision in months (e.g., 18.3 years old). Hence, one unit (0.1) increase in age corresponds to 4.3\% less shooting victimization. For example, 18 years old person is 43\% more likely to be victimized compared to 19 years old person. CIRV list members are four times more likely to be victimized for shooting fatal and non-fatal shootings compared to non-CIRV members. Past misdemeanor\footnote{This finding is somehow interesting; however, from our corresponding author year of experience with the police, data suggest that violent individuals are also violent at home. They commit the majority of domestic violence, and simple assaults that fall under the misdemeanor crime category.} crime history increases the chance of future shootings nearly four times as well compared to individuals having no or less misdemeanor criminal history. 

Finally, recent firearm crime involvement increases the likelihood of future shooting victimizations for 182\%. During the analysis, we noticed that recent violent victimization and firearm-related crimes are moderately correlated (r=0.530). Even though this correlation is under 0.7, it still shadows the effect of recent violent victimization for about 28\% on the outcome variable. For this reason, we removed recent violent victimization from the logistic regression equation but added it to the model because the model creates an additive scale for the positional variables of the model, as explained in Table \ref{table4}. In this way, the proposed model does not lose the 28\% explanatory power of recent violent victimization on the outcome variable. 

\begin{table*}%[ht]
	\caption{Personal variables values in the proposed model}
	\label{table4}
	\setlength\tabcolsep{1mm}
	\begin{tabular}{L|C}
		\hline
		\textbf{Variable}&\textbf{Weight}\\
		\hline
		Age	&7-(Age/10)\\
		CIRV List Member&	1\\
		If the person was recently victimized for a violent crime&	1\\
		If the person was recently involved in a violent crime&	1\\
		If the person involved in any firearm related crimes&	1\\
		If the person recently involved in any firearm related crimes&	1.5\\
		If the person committed more than three misdemeanor crimes&	1\\
		If the person recently committed two misdemeanor crimes&	1\\
		If the person was victimized for three misdemeanor crimes&	1\\
		
		\hline
	\end{tabular}
\end{table*}

The model assigns different weights to different ages to reflect the finding of criminological theories that younger people commit more crimes. For instance, if a person is 18 years old, the age weight will be 5.2. The value of seven (7) is a constant value, which dictates that the influence of age at 70 years old becomes none/zero for predicting future violence. We analyzed the fatal and non-fatal shooting victim data between 2010 and 2017 and noticed that the age formula fits well with the age distribution of the historical data as displayed in Table \ref{table5}. 

\begin{table*}%[ht]
	\caption{Applying The Age Weights to the Fatal and Non-Fatal Shooting Victims (Jan. 1, 2010 - Dec. 31, 2017)}
	\label{table5}
	\setlength\tabcolsep{1mm}
	\centering
	\begin{tabular}{C|C|C}
		\hline
		\textbf{Age Groups}&\textbf{\# of Fatal and Non-Fatal Shooting Victims}&\textbf{Min - Max Weights based on The Formula}\\
		\hline
		
		13 - 17&	244&	5.3 - 5.7\\
		18 - 24&	1222&	4.6 - 5.2\\
		25 - 30&	753&	4 - 4.5\\
		31 - 40&	681&	3 - 3.9\\
		41 - 50&	258&	2 - 2.9\\
		51 - 60&	112&	1 - 1.9\\
		61+&	54&	0 - 0.9\\
		
		\hline
	\end{tabular}
\end{table*}

We purposely give higher weights to juveniles because younger people gradually involve in crimes as they build their criminal careers. Therefore, bringing those younger populations to the attention of law enforcement officials before they commit a serious crime might save lives by intervening in the problems at the right time. 

\subsection{Positional Co-offending Variables}

As noted above, the proposed model expands the friendship network (i.e., co-offending networks) until the third degree\footnote{ First-degree network: one\'s immediate friends; second-degree network: one\'s immediate friends of friends; and third-degree network: one's immediate friends of friends of friends}
to completely capture the direct and indirect impact of peers in developing violent criminal behavior (in our case, fatal and non-fatal shootings). In this context, positional co-offending measures imply the co-offending characteristics of individuals (e.g., occupying a key role) relative to others in the co-offending network. As shown in Table~\ref{table6}, there are six positional variables. The model calculates the network measures for degree centrality and PageRank values after building the co-offending network. In a simple definition, degree centrality refers to number of immediate friends~\cite{borgatti2005centrality, ennett2006peer, granovetter1977strength}.

\begin{table*}%[ht]
	\caption{Personal variables values in the model}
	\label{table6}
	\centering
	\begin{tabular}{lcc}%{|L{45mm}|C{14mm}|C{14mm}|}
		\hline
		&\textbf{Min - Max}&\textbf{Mean - \%}\\
		\hline
		PageRank&	0.15 - 7.55&	0.20 (.272)\\
		Closeness to a high PageRank individual at the first degree network&	 0 - 1&	22.96\%\\
		Closeness to a CIRV member at the first degree network&	 0 - 1&	3.03\%\\
		Closeness to a CIRV member at the second degree network&	 0 - 1&	5.55\%\\
		Closeness to a person involved in a shooting crime or victimization at the first degree&	 0 - 1&	3.93\%\\
		Closeness to a person involved in a shooting crime or victimization at the second degree&	 0 - 1&	6.76\%\\
		\hline
	\end{tabular}
\end{table*}

Google developed PageRank~\cite{avrachenkov2006effect,rogers2002google} values to rank websites for their popularity/importance. PageRank uses two criteria in the calculation: (1) the number of links a page gets and (2) links from important webpages receive more weights. PageRank algorithm is calculated using web graph. For example, the uc.edu web page is considered a node/entity and any hyperlink to this website is considered as edges. The rank and popularity of the uc.edu website are calculated based on the number of links it receives and the number of important websites that uc.edu is linked to. Given this context, the current algorithm (VIPAR scores) emulates the usage of Google PageRank calculation. We first calculate the degree centrality to see the popularity of a person. Then, we calculate the number of different events such as arrest, victimization, and field interviews that a person involved in to assess the repeat victim and repeat offending concepts~\cite{lauritsen1995repeat}. Then, similar to the PageRank algorithm, we give higher weights to those individuals having more connections and more events that connect them to other individuals in the co-offending network. Since the peer influence decreases at the second- and third-degree relationships, we give lesser weights for the events that occur in those degrees.\footnote{Table 7 below display the results similar to previous research that the influence of peers degrades at the second and third degree. Since we could not find any significant influence of third-degree friends on nodes, we did not include their weight in the VIPAR scores. Given this context, VIPAR scores assign half weight to the second-degree relationships.}.  Eqn~\ref{formula1} illustrates this idea with simple terms. In the equation, the model gives fewer weights to degree centrality measure while giving two times the higher weight to the number of events. The product of degree centrality and the number of events is standardized by ten because the highest mode of events is generally around 10. We compared our PageRank values with real PageRank values generated by NodeXL software\footnote{NodeXL is free software and directly works with Microsoft Excel as an extension. More information can be found at this link: https://archive.codeplex.com/?p=nodexl} and noticed that the two values are nearly identical. 
\begin{equation}
\label{formula1}
\scriptsize
\mathrm{network}\ \mathrm{measures}= \frac{(\mathrm{Degree\  Centrality} / 2 )+ \#Events)}{10}
\end{equation}

Following the calculation of network measures, the model calculates positional measures of the network, such as having a shooting friend in the first and second-degree co-offending network and closeness to a CIRV List member. The model searches each person's network (e.g., having a shooting friend at the first-degree co-offending network) to generate the data. If a person is connected to an individual who has a PageRank value greater than one, that person is considered to connect a high PageRank\footnote{If the PageRank value is greater than 1, closeness to a high PageRank individual at the second degree is equal to one otherwise zero.}
individual.  

Likewise, the model searches CIRV member friendship at the first and second-degree co-offending network and assigns a dummy variable code (1 and 0 represent yes and no, respectively) based on the found criteria. Finally, the data include dummy variables by exploring whether a person is connected a shooting victim or suspect at the first and second degree\footnote{We did not include third-degree positional measures because both research (Fujimoto et al., 2012; Papachristos et al., 2015) and my experience suggest that third-degree positional measures/variables have no or little influence on individuals.}.

\begin{table*}%[ht]
	\caption{Logistic Regression Results of Positional Co-offending Variables}
	\label{table7}
	\centering
	\begin{tabular}{lcccc}%{|L{48mm}|C{7mm}|C{7mm}|C{7mm}|C{7mm}|}
		\hline
		&\textbf{b}&	\textbf{S.E.}&	\textbf{p-value}&	\textbf{Exp (B)}\\
		\hline
		PageRank&	0.779&	0.108&	0.000&	2.179\\
		Closeness to a high PageRank individual at the first degree network&	0.752&	0.181&	0.000&	2.121\\
		Closeness to a CIRV member at the first degree network&	0.403&	0.219&	0.066&	1.496\\
		Closeness to a CIRV member at the second degree network&0.135&	0.239&	0.572&	1.144\\
		Closeness to a CIRV member at the third degree network&	0.339&	0.238&	0.155&	1.403\\
		Closeness to a person involved in a shooting crime or victimization at the first degree&	0.609&	0.224&	0.006&	1.839\\
		Closeness to a person involved in a shooting crime or victimization at the third degree&	0.098&	0.241&	0.683&	1.103\\
		\hline
	\end{tabular}
\end{table*}

Given this context, Table~\ref{table7} displays logistic regression results of positional variables. The most influential positional variable in the logistic regression equation is PageRank values. 

Note that the PageRank is a metric variable; therefore, one unit (0.05) increase in the PageRank value corresponds to 2.18 times higher future shooting victimizations. Restating differently with an illustration, a person having a PageRank value two (2) is 43.6 times more likely to be victimized compared to a person having a PageRank value one (1).

On the other hand, if a person is connected to a high PageRank friend at the first-degree co-offending network, that person's vulnerability for future shooting victimizations increases for 212\%. 

Likewise, individuals who have a friend involved in a shooting crime (either suspect or victim) at the first degree co-offending network are 1.8 times more likely to be victimized for future shooting victimizations. 

All the other variables, such as having a CIRV member\footnote{We purposely included CIRV members for all co-offending degrees to better understand the influence of CIRV members in the overall network.} friend at the first, second, and third-degree co-offending network have insignificant influence on the outcome variable. Based on the logistic regression results, the model uses certain weights, as shown in Table~ref{table8}.

\begin{table*}%[ht]
	\caption{Positional Measures of Co-offending Network}
	\label{table8}
	\centering
	\begin{tabular}{lc}%{|L{71mm}|C{13mm}|}
		\hline
		\textbf{Positional Measures}&\textbf{Weight}\\
		\hline
		PageRank value	Own Value
		Closeness to a high PageRank person in first-degree network&	1\\
		Closeness to a CIRV List member in first-degree network&	0.5\\
		If a first-degree friend is involved in any shootings&	1\\
		\hline
	\end{tabular}
\end{table*}

\section{Structural Co-offending Variables}

There are a bunch of studies suggesting violent groups involve in more violence such as in~\cite{battin1998contribution, curry2002gang,gover2009adolescent,huff1998comparing, loeber2001juvenile}. For this reason, the proposed model identifies structural group characteristics such as the number of violent individuals, the number of shootings, and the number of firearm-related incidents in each co-offending group. In order to form co-offending groups, we start with a person and calculate all connected friends until third-degree friendship. This gives us the group of related people. Then, we calculate necessary group-based variables such as the number of individuals involved in violent crimes, shooting, and firearm-related crimes. We loop through this process for all individuals to calculate their co-offending group characteristics. After forming the co-offending groups, we give higher weights to those individuals if they are nested in violent groups.  In this vein, we converted metric variables into categorical variables, as shown in Table \ref{table9} to see the significance level of each group-level variable. There are four group-level variables: violent crime, violent victimization, shootings, and group density.

\begin{table*}%[ht]
	\caption{Descriptive Statistics of Structural Variables}
	\label{table9}
	\centering
	\begin{tabular}{lcc}%{|L{38mm}|C{14mm}|C{14mm}|}
		\hline
		&\textbf{Min - Max}&\textbf{Mean - \%}\\
		\hline
		&Min - Max&	Percentage\\
		Whether a group has more than 3 violent crime&	0 - 1&	10.02\%\\
		Whether a group has more than 3 violent victimizations&	0 - 1&	17.58\%\\
		Whether a group has more than 3 shootings&	0 - 1&	5.46\%\\
		Whether a group has more than 20 members&	0 - 1&	20.12\%\\
		
		\hline
	\end{tabular}
\end{table*}

Moreover, we noticed during the analysis that group-level violent crimes and violent crime victimizations are moderately correlated (r=.653), which then hinders to see their actual effects on the outcome variable. For this reason, the violent victimization variable lost its significance level because of this moderate correlation, as shown in Table \ref{table10}. However, we still added this variable into model prediction because the model creates an additive scale that is not affected by the collinearity.

\begin{table*}%[ht]
	\caption{Logistic Regression Results of Structural Variables}
	\label{table10}
	\centering
	\begin{tabular}{lcccc}%{L{38mm}|C{7mm}|C{7mm}|C{7mm}|C{7mm}}
		\hline
		&	\textbf{b}&	\textbf{S.E.}&	\textbf{p-value}&	\textbf{Exp (B)}\\
		\hline
		Whether a group has more than 3 violent crime&	0.567&	0.193&	0.003&	1.762\\
		Whether a group has more than 3 violent victimizations&	0.333&	0.212&	0.117&	1.395\\
		Whether a group has more than 3 shootings&	0.792&	0.183&	0.000&	2.207\\
		Whether a group has more than 20 members&	0.886&	0.204&	0.000&	2.426\\
		
		\hline
	\end{tabular}
\end{table*}

Table~\ref{table10} suggests that individuals nested in populated and shooting dense groups are nearly two times higher vulnerability for future shooting victimizations compared to less populated and less shooting dense groups. Likewise, members of violent groups are more likely to be a victim of future shootings. Given the findings of structural variables, the model incorporates slightly different group-level variables, as seen in Table~\ref{table11} to fully reflect variable variations into the prediction.  

\begin{table*}%[ht]%[h]
	\caption{Structural Measures of Co-Offending Network}
	\label{table11}
	\center
	\begin{tabular}{lc}%{L{45mm}|C{35mm}}
		\hline
		\textbf{Structural Measures}&\textbf{Weight}\\
		\hline
		Number of violent crime in the co-offending network&	>=10= 3; between 5 and 9=2; between 2 and 5=1\\
		Number of violent victimizations in the co-offending network&	>=10= 3; between 5 and 9=2; between 2 and 5=1\\
		Number of recent violent victimizations in the co-offending network&	>=7=2\\
		Number of recent shootings in the co-offending network&	>=1=2\\
		Number of shootings in the co-offending network&	>=3=1\\
		If the co-offending group has more than 20 members&	1\\
		If the co-offending group has more than 10 shootings&	1\\
		If the co-offending group has more than 5 recent shootings&	1\\

		\hline
	\end{tabular}
\end{table*}

\section{Summary of the Measures} 

The proposed model (VIPAR scores) firstly calculates age, group membership status (e.g., CIRV list affiliation\footnote{Being a CIRV list member can also be considered as a personal characteristic (e.g., age) rather than a positional measure.}), and past criminal history of individuals. Next, it computes positional measures (e.g., occupying a key role in the network) and structural measures (e.g., being a member of a violent group) of individuals in the co-offending network. Positional measures are generally related to individuals' positions in the co-offending network relative to others. For instance, group\footnote{According to focused deterrence approach, the vast majority of violence in any city is committed by a small group of offenders, who are connected socially in groups~\cite{kennedy1996youth,peterson2004gang}. These groups do not always adhere to the traditional hierarchy associated with gangs and are mostly comprised of loose-knit social networks of individuals that offend together. For this reason, rather than calling gangs, the researcher tends to rename these socially connected people as group members. 
	
As stated earlier, the City of Cincinnati implemented a focused deterrence approach called CIRV. As a result of intelligence gathering sessions, field experts identified CIRV group members. These individuals and groups are known for their violence propensities in the city. Therefore, the present model also uses this export knowledge input and gives higher weights to individuals close to those groups members (called as CIRV list in this paper).} members receive higher weights, which is aligned with the findings of focused deterrence approach. 

Likewise, the model calculates the precise PageRank value of each person and assigns higher weights to high PageRank individuals. The remaining three positional measures are all shaped by the positions of individuals in the co-offending network. For instance, if a person's\footnote{it is called as ego in network graph theory} friend\footnote{it is called as an edge in network graph theory} involves in a shooting crime, that person gets weight for having a shooting friend in the first degree co-offending network. Finally, since the main aim of this model is to identify who is likely to commit fatal and non-fatal shooting crimes or being a victim of a shooting crime, the model gives higher weights to those individuals if their friends are involved in either shootings or violent crimes.

Structural measures are primarily related to the characteristics of groups in which individuals are nested. For instance, if a person is nested in a violent co-offending group, that person receives a higher weight. Likewise, the proposed model gives higher weights to individuals if the number of fatal and non-fatal shootings and firearm-related crimes is high in the group. As a result of the computing process, the model generated VIPAR scores for 55,454 individuals ranging from 1.05 to 28.45

To summarize, the designed model takes into account the age of individuals, past criminal history, and peer influence using the principles of network graph theory. Note that the model does not include any gender, race, and place characteristics for racial profiling concerns. In crime prevention theory, place characteristics (e.g., risky places such as bars) are good predictors of future crime concentration~\cite{brantingham1984patterns}. Certain areas, however, predominantly contain a specific racial group, therefore, including place-based characteristics may lead to hidden racial bias in the model. Due to this concern, the model excludes place-based characteristics.

\section{Results}
\begin{figure}
	\centering
	\includegraphics[width=8cm, height=5cm]{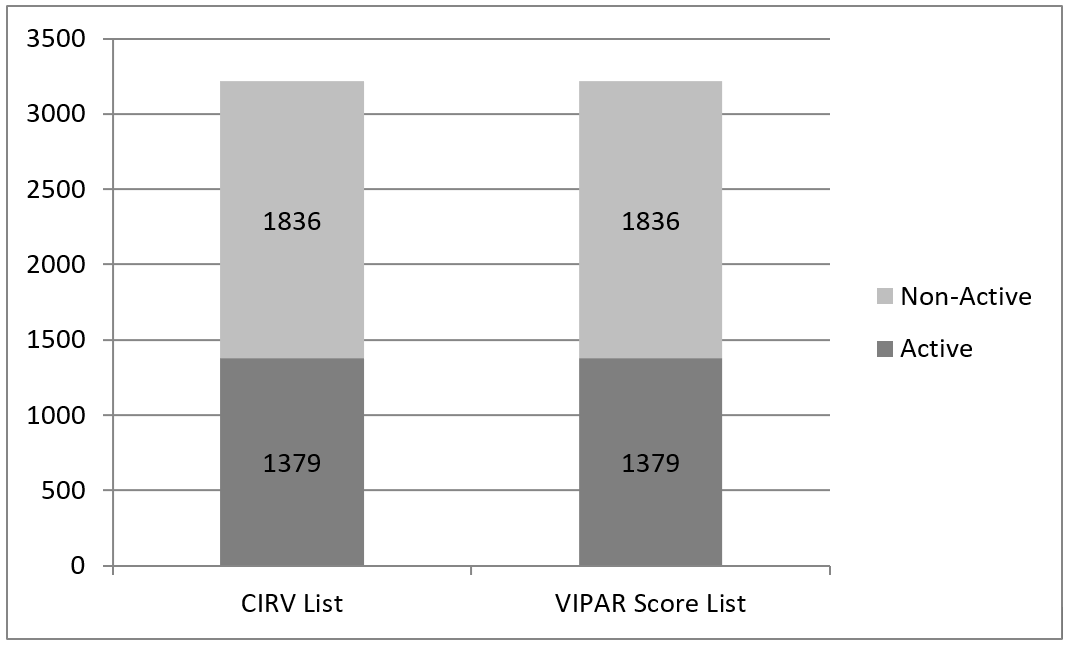}
	\caption{CIRV List and VIPAR Score List}
	\label{Fig:fig3}
\end{figure}

As previously stated, the VIPAR score model employed the data from July 1, 2010, to December 31, 2014, to rank individuals for their vulnerability regarding future fatal and non-fatal shootings. Given this context, the model generated VIPAR scores for 55,454 individuals. In addition to VIPAR scores, Cincinnati has CIRV list\footnote{Note that 2014-2015 CIRV list was used to predict 2015 shootings.}
to predict future fatal and non-fatal shootings. The CIRV list contains 1,379 active key players and 1,836 non-active members. The second research question of the study was to find out whether VIPAR scores better predict future shootings than the CIRV list.
For this reason, as Figure~\ref{Fig:fig3} displays, the present study employed the first top 1,379 individuals in the VIPAR score list as active members, and the next 1,876 as non-active members to have an equal number of cases in both samples (VIPAR score list and CIRV list) for a fair comparison. 

Following adjusting the two samples, the study first compared the two lists (CIRV and VIPAR) for their prediction power regarding future fatal and non-fatal shooting suspects. The proposed model requires to have full names and date of births' of future suspects to make a comparison between the VIPAR scores and the CIRV list. According to 2015 statistics, Cincinnati Police Departments could identify 149 suspects out of 477 fatal and non-fatal shootings by their full name and date of birth\footnote{To see the matched names among samples, full name, and date of birth fields are necessary}. Upon identification of 2015 suspects, the study matched the VIPAR score list and the CIRV list with the known 2015 shooting suspects using full names and date of births. Figure~\ref{Fig:fig4} below displays the result of this matching procedure. Results show that the first top 1,379 VIPAR score list predicts 34 out of 149 (22.8\%) 2015 shooting suspects, and the second top 1,836 VIPAR score list predicts 14 out of 149 (9.4\%). The entire VIPAR score list (N=3,215) predicts 32.2\% of known shooting suspects for the year of 2015. In other words, the VIPAR score model successfully predicts nearly one-third of the future shooting suspects. On the other hand, active CIRV List and non-active CIRV list predict 9.4\% of the future known shooting suspects.

\begin{figure}
	\centering
	\includegraphics[width=8cm, height=5cm]{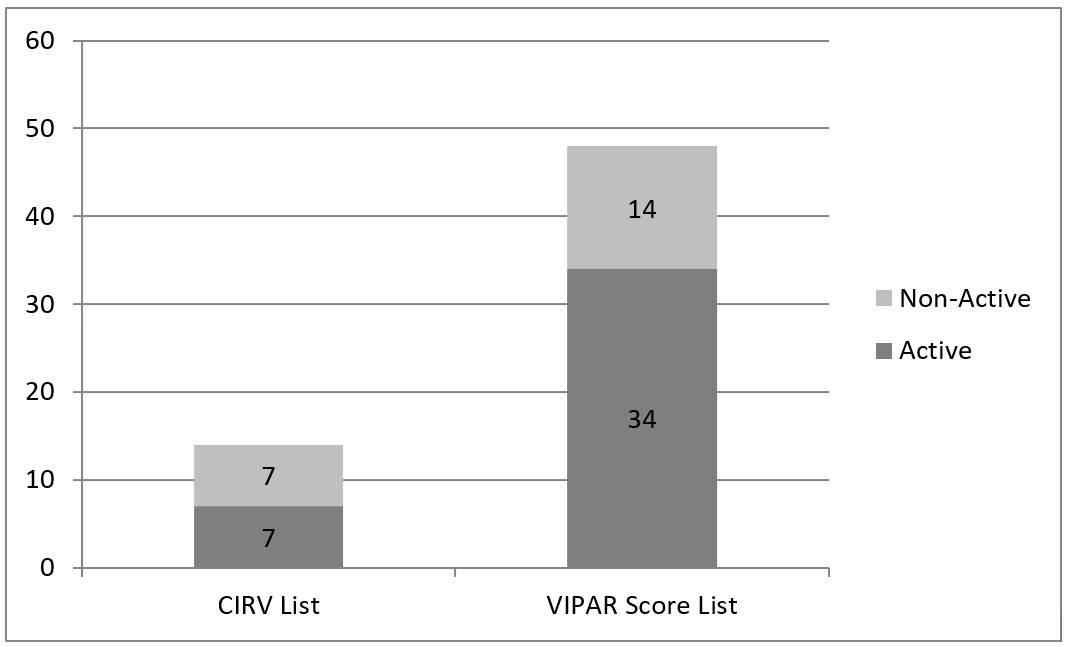}
	\caption{Predicting Future Fatal and Non-Fatal Shooting Suspects (N=149)}
	\label{Fig:fig4}
\end{figure}

\begin{figure}
	\centering
	\includegraphics[width=8cm, height=5cm]{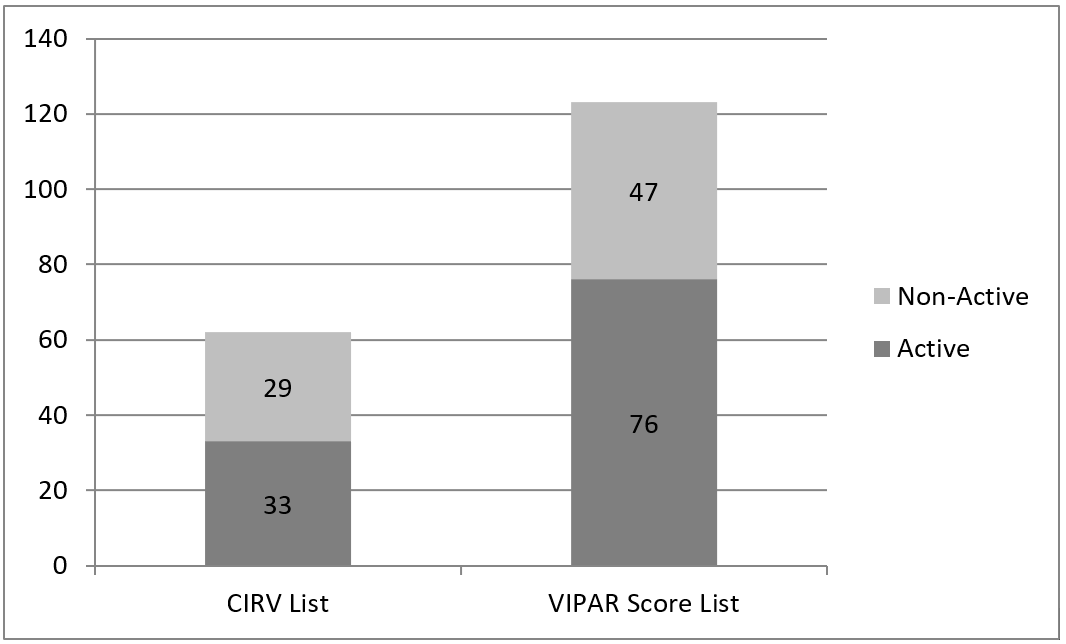}
	\caption{Predicting Future Fatal and Non-Fatal Shooting Victims (N=477)}
	\label{Fig:fig5}
\end{figure}

The second set of analyses includes the prediction of future shooting victims that occurred in 2015. As seen in Figure~\ref{Fig:fig5}, the VIPAR score list predicts 123 out of 477 (25.8\%) shooting victims. In other words, VIPAR scores successfully predicted every 1 out of 4 shooting victims. On the other hand, the CIRV list predicted 13\% of the future shooting victims in the city. These primitive comparisons suggest \footnote{The entire VIPAR list was including 55,544 individuals when we conducted the analysis. Therefore, we used top 5.8 percent of the VIPAR list to evaluate the predictive success of VIPAR list.}.

\section{Discussion and Conclusion}
There is a growing concern that prioritizing individuals for their vulnerability scores might violate the presumption of innocence right of people~\cite{asher2017inside}. The present study fully shares the same concerns for any computerized intelligence if the model is not based on scientific theories and also not publicly available upon request. Given this context, criminology is a well-established discipline that has century-old studies and theories. As a rule of research methodology, once the variable relationships (e.g., age and crime) are confirmed from one study to another, researchers tend to believe that the magnitude of the relationship is causal or almost a causal relationship. As stated at the beginning of this paper, the criminological studies repeatedly suggest that age, past criminal history, and peer influence are the most important predictors of future delinquent behavior. Therefore, following the pure thoughts/science of the criminology field during the development of any crime-related model might mitigate the current liberty concerns. 

Furthermore, variables and methods used in any crime related model should be public to share the science behind computerized intelligence. The other balance check method is that experts should always validate the computer information before making any decision for intervention. Finally, the model should avoid to include any variables (e.g., race, gender, and place-based characteristics) that might lead to profiling. Certain places predominantly contain specific racial groups; therefore, even including place-based characteristics might lead to indirect racial profiling when developing a model to predict vulnerable populations. 

Given this context, VIPAR scores systematically analyze the large volume of data with automated codes to predict future shooting victims and suspects. The model is open and only uses the proven thoughts of criminological theories. VIPAR scores aim to identify emerging vulnerable populations, specifically juveniles, to take actions on time to save lives. As an important note, even though VIPAR scores successfully identify vulnerable populations, law enforcement officials should not solely use it for aggressive style policing by targeting top-ranked individuals. 

Focused deterrence strategies can be a practical implementation of VIPAR scores. The proposed model partly employs the CIRV list, which was generated within the principles of a focused deterrence approach. Findings suggest that VIPAR scores better predict the future shooing violence than the CIRV list after adding relevant variables (e.g., age, past criminal history, peer influence) from criminological theories. In this context, VIPAR scores can be a component of focused deterrence policing, which is known as an effective way to reduce violence in cities~\cite{braga2012effectsb, engel2013reducing, nagin2013deterrence}.

There are certain limitations to the development of VIPAR scores. First, the data used in the algorithm is police-reported incidents; therefore, the algorithm will be limited to those individuals who have records/contacts with the police. Even though this is a limitation of the VIPAR scores, it is also a strength because the algorithm only uses the police contacts rather than using any source of subjective data (e.g., social media). The other likely limitation is that as explained in the VIPAR scores, PageRank values are calculated based on the number of immediate friends and the number of different events. (e.g., arrest, field interviews). Readers might ask that the more targeted police contacts, specifically through field interviews, the higher the VIPAR scores. This may not be a concern for VIPAR scores because the number of police contacts has an ignorable effect in the calculation. The main components of VIPAR scores are age, violence, and peer influence. Therefore, frequent police contacts will not increase a person’s position in the VIPAR score list unless that person involves in violent crime and has prolific friends at the first-degree relationship. Please note that we developed our based on criminological theories; therefore, VIPAR scores only converts the proven thoughts of criminology into digits in order to prioritize proactive efforts of police departments to effectively save lives. Having said that, VIPAR scores should be periodically checked by researchers in case of any possible bias which might substantially affect the overall purpose of the system.   

\printcredits


\bibliographystyle{cas-model2-names}

\bibliography{main}

\end{document}